\pdfoutput=1
\def \isblinded {no}
\documentclass[11pt]{article}

\def \toblind {yes}
\ifx \isblinded \toblind
	\usepackage[review]{acl}
\else
	\usepackage[final]{acl}
\fi

\usepackage{times}
\usepackage{latexsym}

\usepackage[T1]{fontenc}

\usepackage[utf8]{inputenc}

\usepackage{microtype}

\usepackage{inconsolata}

\usepackage{graphicx}
\usepackage{booktabs}
\usepackage{color}

\title{A Two-dimensional Zero-shot Dialogue State Tracking Evaluation Method using GPT-4}

\author{Ming Gu \and Yan Yang\thanks{Corresponding Author} \\
	School of Computer Science and Technology, East China Normal University \\
	\texttt{51215901012@stu.ecnu.edu.cn}, \texttt{yanyang@cs.ecnu.edu.cn}}

\begin{document}
\maketitle
\begin{abstract}
Dialogue state tracking (DST) is evaluated by exact matching methods, which rely on large amounts of labeled data and ignore semantic consistency, leading to over-evaluation. Currently, leveraging large language models (LLM) in evaluating natural language processing tasks has achieved promising results. However, using LLM for DST evaluation is still under explored. In this paper, we propose a two-dimensional zero-shot evaluation method for DST using GPT-4, which divides the evaluation into two dimensions: accuracy and completeness. Furthermore, we also design two manual reasoning paths in prompting to further improve the accuracy of evaluation. Experimental results show that our method achieves better performance compared to the baselines, and is consistent with traditional exact matching based methods.
\ifx \isblinded \toblind
\else
	\footnote{The code is available at \url{https://github.com/SLEEPWALKERG/LLM-DST-EVAL}}
\fi
\end{abstract}

\section{Introduction}
Dialogue state tracking (DST) is a key component in task-oriented dialogue systems, aiming at tracking all key information during a dialogue. The primary metrics of this task include joint goal accuracy and co-reference slot accuracy, which compare the predicted state with the ground truth. All of these metrics rely on large amounts of annotated data. In practice, due to the quick emergence of new domains and the high cost associated with the data annotation, an evaluation method that does not require annotated data is urgently needed for DST.

Furthermore, there is an inherent flaw in current evaluation methods that they compare the predicted state with the ground truth in an exact match manner, leading to over-evaluation problems. DST is a task of natural language understanding, so it is unreasonable to simply evaluate it by string match. The over-evaluation phenomenon is notably prevalent in the MultiWOZ dataset as shown in Table \ref{examples-of-over-evaluation}. In the first example, the model copies the "pizza hut fen ditton" from the context, and it is correct. However, it cannot be matched to the annotation. The value in the second example contains an additional definite article "the", but the meaning is correct. More attention should be paid to developing a more reasonable evaluation method for DST.
\begin{table}[t]
	\centering
	\small
	\begin{tabular}{p{0.95\columnwidth}}
		\toprule
		\textbf{Dialogue Turn:} [sys] ... [user] I would like a taxi from ... to pizza hut {\color{blue}{fen ditton}}.\\
		\textbf{Model Output:}\\taxi-destination: pizza hut {\color{blue}{fen ditton}}, ...\\
		\textbf{Ground Truth:}\\taxi-destination: pizza hut {\color{red}{fenditton}}, ...\\
		\midrule
		\textbf{Dialogue Turn:} [sys] What is the name of the hotel ... [user] The gonville. Have you heard of it?\\
		\textbf{Model Output:}\\hotel-name: {\color{blue}{the}} gonville hotel\\
		\textbf{Ground Truth:}\\hotel-name: gonville hotel\\
		\bottomrule
	\end{tabular}
	\caption{Two Examples of over-evaluation caused by exact matching evaluation.}
	\label{examples-of-over-evaluation}
\end{table}

Recently, large language models (LLM) have shown promising performance in evaluating different natural language processing tasks\cite{gptscore, chatgpt-as-evaluator, gemba, llm-as-explainable-metric, instructscore, gpteval}. Evaluating DST models with LLM can not only get rid of the dependence on labeled data but also mitigate the over-evaluation problem. However, most of these research points to the natural language generation (NLG) tasks. Different from NLG, the information extraction (IE) task has stricter restrictions on the expression and the direction of evaluation varies greatly. Leveraging LLMs for evaluating generative IE tasks like DST is still under explored.
\begin{figure*}[ht]
	\begin{center}
		\includegraphics[width=0.9\textwidth]{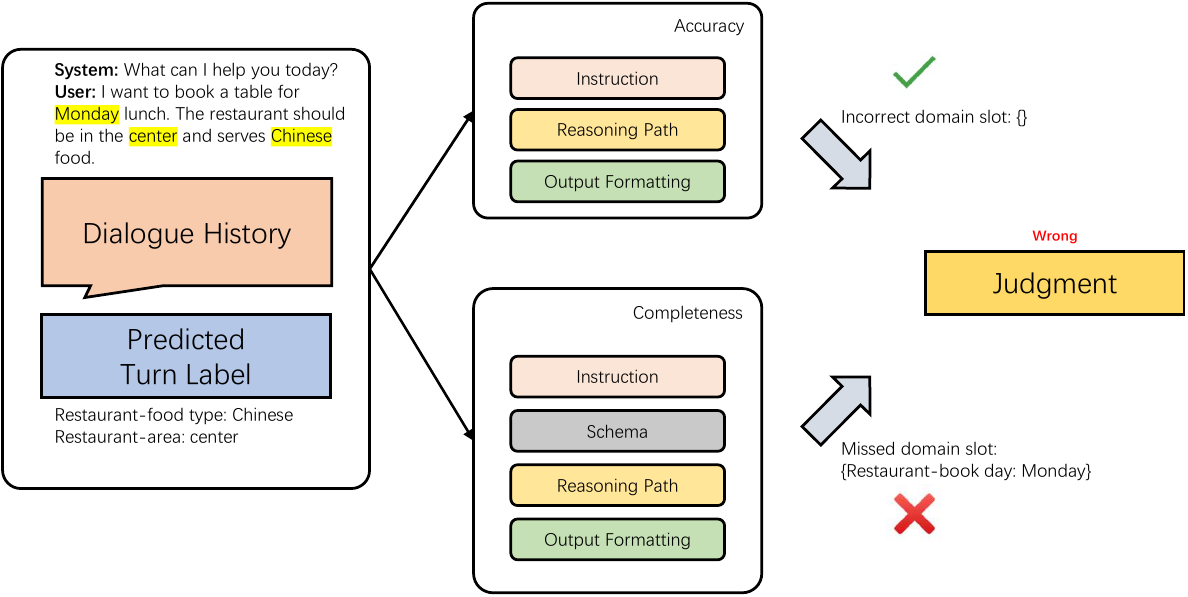}
	\end{center}
	\caption{The framework of our proposed evaluation method for dialogue state tracking.}
	\label{framework}
\end{figure*}

In this paper, we propose a two-dimensional zero-shot evaluation method for DST, which does not require any annotated data with the help of LLMs' strong capability of reasoning. Specifically, we first divide the evaluation into two dimensions: accuracy and completeness, aiming at more fine-grained evaluation. Moreover, we design two manually constructed reasoning paths for the two prompt templates to help the LLM better understand the emphasis of the two dimensions, aiming at more accurate evaluation. Experimental results show that our proposed evaluation method achieves an assessment accuracy of 91\% for turn state accuracy, which is comparable to the match-based method's 95\%, and maintains consistency, offering a general zero-shot evaluation scheme for dialogue state tracking. 

The contributions of this paper are summarized as the following:
\begin{itemize}
	\itemsep0em
	\item We propose a two-dimensional zero-shot evaluation method for DST.
	\item We design manual reasoning paths for evaluation of both accuracy and completeness to further improve the evaluation accuracy.
	\item Experimental results show the superiority of our proposed method for evaluating the DST models' performance. 
\end{itemize}

\section{Methodology}
In this section, we will describe our two-dimensional framework. Figure \ref{framework} illustrates the whole process of our method. We divide the evaluation into two dimensions: accuracy and completeness. Only if the judgment of both dimensions is correct, the turn state is considered correct. 
\subsection{Accuracy Evaluation}
For the correctness of the predicted state, the accuracy is crucial. Once an error occurs, the resulting accumulation of errors can be fatal. We carefully construct a manual prompt template to instruct the LLM to judge whether all \{domain-slot, value\} pairs in the predicted turn state are accurate. For better judgment, we also manually construct a reasoning path to help the LLM better understand the key point of the accuracy evaluation. We ask the model to read the predicted turn state one by one and judge the accuracy of each \{domain-slot, value\} pair by referring to the context. Finally, we also specify the output format of the LLM, which is in JSON format like \{"explanation": <the logical progression of reasoning>, "incorrect\_domain\_slots", <all incorrect domain-slots in the predicted turn state>\}.
\subsection{Completeness Evaluation}
The completeness of the predicted turn state is another important point for its correctness because the turn state should catch all the requirements and confirmation of the user in a dialogue turn. It is a big challenge for the LLM to asses the completeness since it requires the LLM to have a deep understanding of the schema and dialogue, and then reason to determine the completeness. The constructed prompt template for this dimension consists of instruction, schema, reasoning path, and the output format. Contrary to the reasoning path designed for accuracy assessment, The reasoning path for evaluating completeness asks the LLM to first read the turn utterances. Then, when encountering a slot value raised or confirmed by the user, we instruct the LLM to judge whether it is in the predicted turn state. Finally, the LLM is asked to give the final result based on the aforementioned procedure. The output format for this dimension is also set to JSON like in accuracy evaluation. Notably, we observe that the LLM occasionally appends some domain-slots that have already been accounted in the previous turn or even not in the schema. To address this, we construct some rules to filter out these errors.
\subsection{Result Integration}
After obtaining all missed \{domain-slot, value\} pairs and the incorrect ones within a given turn state, it is easy to get the turn state accuracy. If and only if both the missed and incorrect \{domain-slot, value\} pairs lists are empty, the turn state is considered correct. Note that we further maintain a list of correct states to save all \{domain-slot, value\} pairs that have been judged correct, and then if the predicted turn state contains \{domain-slot, value\} pairs in this list, it will be treated as incorrect.

Furthermore, it is also easy to gain the joint goal accuracy (JGA) score, which is the the standard metric in DST\cite{trade}, because the tracking of dialogue states is an inherently progressive process updated via turn states. To gain the JGA score, we maintain two lists of already incorrect states and already missed states. If and only if both of the two lists is empty, then the dialogue state is considered correct. At each dialogue turn, we use the missed, incorrect, and correct \{domain-slot, value\} pairs to update these two lists.
\section{Experiments}
\subsection{Datasets and Metrics}
\textbf{Datasets} We choose the data from SVAG\cite{svag} and EDZ-DA\cite{edz-da}. SVAG provides its test results on MultiWOZ\cite{mwz2-1}. EDZ-DA is a data augmentation method for task-oriented dialogue and EDZ-DA provides the test results of the augmented SVAG. Since almost all kinds of errors can be found in extreme low-resource scenarios, we use their data that the model is trained under the data ratio setting of 1\% to evaluate whether our proposed method can catch all these errors. We sample 100 dialogues from the test set for evaluation.

\textbf{Metrics} We adopt the accuracy of Turn State Accuracy (TSA) to evaluate the performance of different evaluation methods. We use the test set of MultiWOZ 2.4\cite{mwz2-4} to verify the accuracy of different evaluation methods. Furthermore, we manually check samples that the evaluation of MultiWOZ2.4 and our method is inconsistent. The reason why we do not evaluate the accuracy of the standard metric for DST, joint goal accuracy (JGA) is that it will lead to evaluation bias since the dialogue state is a process of continuous update by the turn state, which is a cumulative process in most cases. For example, there is a pair of \{domain-slot, value\} appearing in the first dialogue turn, and it is not changed in the following dialogue. Then evaluation based on JGA will evaluate this \{domain-slot, value\} pair n times where n is the turns of the dialogue. If the judgment of this \{domain-slot, value\} pair is different among different evaluation methods, the evaluation bias occurs.
\subsection{Experimental Settings}
We employ the GPT-4 Turbo model available in OpenAI API\footnote{\url{https://openai.com}} for our method. In terms of the parameter setting, we set the temperature to 0 and top-p to 1.
\subsection{Baselines}
We compare our method with the following baselines:

\textbf{Direct} instructs the LLM to directly judge the accuracy of the turn state and give some explanation.

\textbf{CoT} adds "Let's think step by step" to the direct prompt.

\textbf{Two-dimensional CoT} adds "Let's think step by step" to both basic prompt for accuracy and completeness evaluation.
\subsection{Main Result}
\begin{table}[t]
	\centering
	\small
	\begin{tabular}{l|c}
		\toprule
		\textbf{Method} 					& \textbf{Accuracy}   \\
		\midrule
		Direct                       		& 78.42 \\
		\midrule
		CoT                          & 82.10 \\
		\midrule
		Two-dimensional CoT          & 82.92  \\
		\midrule
		Ours                                & \textbf{85.66} \\
		\bottomrule
	\end{tabular}
	\caption{The evaluation accuracy of different methods based on MultiWOZ 2.4.}
	\label{m24-result}
\end{table}
\begin{table}[t]
	\centering
	\small
	\begin{tabular}{l|c}
		\toprule
		\textbf{Method}                    & \textbf{Accuracy}   \\
		\midrule
		Direct                             & 82.79 \\
		\midrule
		CoT                         & 86.34 \\
		\midrule
		Two-dimensional CoT         & 87.3  \\
		\midrule
		Ours                               & 90.85 \\
		\midrule
		RULE-M24                           & \textbf{94.81} \\
		\bottomrule
	\end{tabular}
	\caption{The evaluation accuracy of different methods by human evaluation. "RULE-M24" means the string-match-based evaluation according to the MultiWOZ 2.4 annotation}
	\label{human-result}
\end{table}
Table \ref{m24-result} shows the evaluation accuracy of different methods based on the MultiWOZ 2.4 annotation. For samples where our method is inconsistent with the MultiWOZ 2.4 annotation, we do a further human evaluation and Table \ref{human-result} shows the results. Our proposed method achieves SOTA performance among all strong baselines. Compared to the CoT method that directly gains the judgment, the two-dimensional CoT assessment achieves better evaluation accuracy. Dividing the evaluation into accuracy and completeness allows the LLM to have different focuses in different dimensions and also reduces the difficulty of the task, enabling accurate evaluation in each dimension. Compared to the Two-dimensional CoT method, our proposed method achieves better performance with a great margin, which demonstrates the efficiency of the manual reasoning path designed by us. The manual reasoning path can help the LLM better understand the evaluation direction and focus of different dimensions, leading to more accurate evaluation.

\begin{figure}[t]
	\begin{center}
		\includegraphics[width=0.4\textwidth]{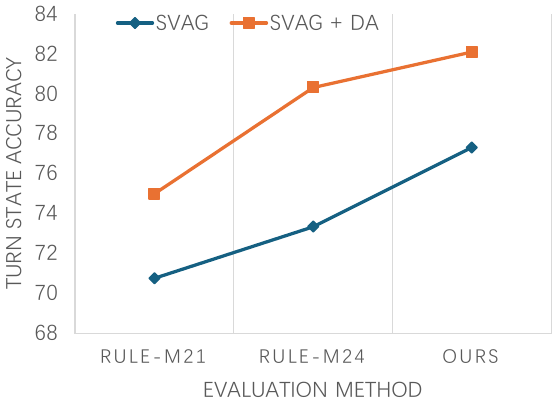}
	\end{center}
	\caption{Turn state accuracy evaluated by different methods.}
	\label{trend-tsa}
\end{figure}

\begin{figure}[t]
	\begin{center}
		\includegraphics[width=0.4\textwidth]{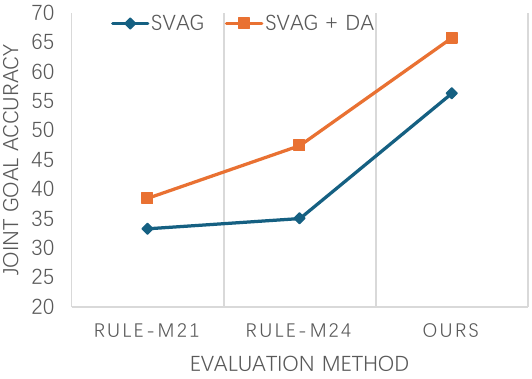}
	\end{center}
	\caption{Joint goal accuracy evaluated by different methods.}
	\label{trend-jga}
\end{figure}

The most crucial function of evaluation methods is to be able to distinguish the performance differences among different models. Therefore, we further evaluate the output of SVAG that has been augmented by EDZ-DA. Figure \ref{trend-tsa} and \ref{trend-jga} show the TSA and JGA score of different models with different evaluation methods. Compared to evaluation methods based on MultiWOZ 2.1 \& 2.4 (RULE-M21 and RULE-M24), the results evaluated by our proposed method are higher since our method gets rid of the over-evaluation problem caused by string-match-based method. Additionally, we observe that our method can also distinguish the performance of different models, and is consistent with previous methods. The results further prove the effectiveness of our method for evaluating DST models.
\section{Related Work}
The primary evaluation metric for DST is joint goal accuracy\cite{trade}. However, this kind of methods highly depends on labeled data and ignores the semantic consistency. Currently, LLM-based evaluation methods\cite{gptscore, chatgpt-as-evaluator, gemba, llm-as-explainable-metric, instructscore, gpteval} have attracted increasing attention and have achieved promising results. Some of these works explore the potential of LLMs to be an explainable metric. However, most of them focus on NLG. Evaluating generative IE models like DST with LLMs is a direction worth studying.
\section{Conclusion}
In this paper, we propose a two-dimensional zero-shot evaluation method for dialogue state tracking using GPT-4. We divide the evaluation into accuracy and completeness and we design reasoning paths for each dimension to help the LLM better understand the focus and direction of different dimensions. Experimental results illustrate the superiority of our proposed method.

In future work, we will adopt more prompting techniques to further improve the accuracy of evaluation.
\section*{Limitations}
In this section, we discuss several limitations of our proposed LLM-based evaluation method for DST. First, it is interesting to adopt diverse prompting techniques to further improve the evaluation accuracy. Second, the prompt template in our paper is manually constructed. Future work can look into automatically constructing the template. Finally, using a smaller LLM like LLaMA\cite{llama} as the backbone model can be further studied. With such a smaller and open-source model, we can also fine-tune it to further investigate the efficiency of a fine-tuned model for DST evaluation.
\section*{Ethical Considerations}
We conduct our experiments upon GPT-4, which is a generative model. So, we carefully check the output. We do not find any harmful information. Furthermore, the dataset we used in this paper is open source. In summary, there are no direct ethical concerns in our study.

%

\bibliography{custom}

\newpage
\appendix
\section{Appendix: Case Study}
\label{appendix-case}
In this section, we give some example output of our proposed evaluation method. Table \ref{case-different-mwz24} shows two examples that the evaluation results are inconsistent with the MultiWOZ 2.4 annotation but are the same as human evaluation. In the first example, our method can evaluate the output from the perspective of natural language understanding (NLU), ignoring the influence of definite articles, and finally give the accurate evaluation. In the second example, the user accepts the system's recommendation, so the ground truth state should include "attraction-name: all saints church", but the annotation of MultiWOZ 2.4 misses it. Besides, our proposed method catches the missing information and gives the correct evaluation. After observing the entire dataset, most of the places that the user confirms are correctly labeled. So, there is still some inconsistency in annotation. For data annotated by multiple annotators, consistency is difficult to effectively guarantee. Therefore, how to use large language models for consistent labeling and evaluation will be a direction worth studying.

Furthermore, we also give three examples that the evaluation of our method are inconsistent with human evaluation as shown in Table \ref{case-different-human}. In the first example, the LLM considers that the exact time for the Taxi should be "after 17:15", which is correct according to the context. However, it is not consistent with the settings in MultiWOZ. Additionally, in the second example, the user exactly expressed that he/she need a "0 star" hotel and the DST model has caught it. The LLM evaluates it as incorrect since a 0 star hotel is not possible in practice. Both the above cases show that the LLM has its own set of strict evaluation rules based on conversational understanding and common sense knowledge. However, sometimes it is excessive. Future work can look into further improving the prompt template to help the LLM better understand the judgment criteria of a specific system such as MultiWOZ. In the last example, the LLM judges "restaurant-pricerange: high" is accurate, but ignores that the value of "pricerange" should be in \{"cheap", "moderate", "expensive"\}, which is specified in the schema. In future work, to avoid this kind of error, we can add more constraints to the prompt.
\begin{table}[t]
	\centering
	\small
	\begin{tabular}{p{0.95\columnwidth}}
		\toprule
		\textbf{Dialogue history:} None\\
		\textbf{Current Turn Utterances:} [sys] ... [user] I need info about {\color[RGB]{255, 0, 0}{the slug and lettuce restaurant}}. \\
		\textbf{Model output:} restaurant-name: slug and lettuce \\
		\textbf{MultiWOZ 2.4 Annotation:} restaurant-name: {\color[RGB]{255, 0, 0}{the}} slug and lettuce \\
		\textbf{MultiWOZ 2.4 Evaluation:} \color{red}{Incorrect} \\
		\textbf{Evaluation of ours:} \color[RGB]{84, 180, 53}{Correct} \\
		\textbf{Human Evaluation:} \color[RGB]{84, 180, 53}{Correct} \\
		\textbf{Evaluation Explanation of ours:} The user has mentioned `slug and lettuce restaurant' which indicates they are looking for information about a specific restaurant named Slug and Lettuce. The `restaurant-name' slot has been correctly identified with the value `slug and lettuce'. There is no indication in the user's request that the information provided in the `State of this turn' is incorrect. \\
		\midrule
		\textbf{Dialogue history:} [sys] booking is successful. ... [user] Can you suggest any places to go in town? ... [user] How about architecture? I will need the entrance fee, phone number, and postcode. \\
		\textbf{Current Turn Utterances:} [sys] I would recommend {\color[RGB]{255, 0, 0}{all saints church}}. The entrance fee is free, the postcode is cb58bs, and the phone number is 01223452587. [user] Great. That is all the info I needed today. \\
		\textbf{Model output:} None\\
		\textbf{MultiWOZ 2.4 Annotation:} None \\
		\textbf{MultiWOZ 2.4 Evaluation:} \color[RGB]{84, 180, 53}{Correct} \\
		\textbf{Evaluation of ours:} \color{red}{Incorrect} \\
		\textbf{Human Evaluation:} \color{red}{Incorrect} \\
		\textbf{Evaluation Explanation of ours:} In the current turn, the agent provides a recommendation for an attraction, which is All Saints Church. The agent also provides the entrance fee, postcode, and phone number for the attraction. Since the user confirms that this is all the information needed, it is implied that the user accepts the recommendation. Therefore, the turn state should include the details of the recommendation. ... \\
		\bottomrule
	\end{tabular}
	\caption{Two examples that the evaluation results of ours are coherent to the human evaluation but different from the MultiWOZ 2.4 annotation.}
	\label{case-different-mwz24}
\end{table}
\begin{table*}[t]
	\centering
	\small
	\resizebox{0.98\linewidth}{!}{
			\begin{tabular}{p{0.95\linewidth}}
				\toprule
				\textbf{Dialogue history:} ... [user] I would like a taxi from saint john s college to pizza hut fen ditton. \\
				\textbf{Current Turn Utterances:} [sys] What time do you want to leave and what time do you want to arrive by? [user] I want to leave after {\color[RGB]{84, 180, 53}{17:15}}. \\
				\textbf{Model output:} Taxi-leaveat: 17:15 \\
				\textbf{MultiWOZ 2.4 Annotation:} Taxi-leaveat: 17:15 \\
				\textbf{MultiWOZ 2.4 Evaluation:} \color[RGB]{84, 180, 53}{Correct} \\
				\textbf{Evaluation of ours:} \color{red}{Incorrect} \\
				\textbf{Human Evaluation:} \color[RGB]{84, 180, 53}{Correct} \\
				\textbf{Evaluation Explanation of ours:} The user has specified that they want to leave `after 17:15'. The state captures the time as `17:15', which is not entirely accurate because `after 17:15' implies any time that is later than 17:15, not exactly at 17:15. Therefore, the value should reflect that the departure time is not strictly 17:15 but some time after that. \\
				\midrule
				\textbf{Dialogue history:} ... [sys] Certainly, what sort of hotel did you have in mind. [user] I would kile a hotel in the moderate price range with free wifi. \\
				\textbf{Current Turn Utterances:} [sys] I have 17 hotels that fit those criteria, is there a particular area or type of accommodation you would prefer? [user] What have you got with {\color[RGB]{84, 180, 53}{0 stars}}? \\
				\textbf{Model output:}	hotel-stars: 0\\
				\textbf{MultiWOZ 2.4 Annotation:} hotel-stars: 0 \\
				\textbf{MultiWOZ 2.4 Evaluation:} \color[RGB]{84, 180, 53}{Correct} \\
				\textbf{Evaluation of ours:} \color{red}{Incorrect} \\
				\textbf{Human Evaluation:} \color[RGB]{84, 180, 53}{Correct} \\
				\textbf{Evaluation Explanation of ours:} The user asked for hotels with `0 stars', which is an unusual request as hotel star ratings typically start at 1 star. ... 
				Since the user's intent is not entirely clear and could be interpreted in different ways, it is difficult to definitively categorize the `0 stars' request as correct or incorrect without further clarification from the user.\\
				\midrule
				\textbf{Dialogue history:} None \\
				\textbf{Current Turn Utterances:} [sys] ... [user] I am looking for a {\color[RGB]{84, 180, 53}{high end indian}} restaurant, are there any in town ? \\
				\textbf{Model output:} restaurant-pricerange: high, restaurant-food: indian \\
				\textbf{MultiWOZ 2.4 Annotation:} restaurant-pricerange: high, restaurant-food: indian, {\color[RGB]{255, 0, 0}{restaurant-area: dontcare}} \\
				\textbf{MultiWOZ 2.4 Evaluation:} \color{red}{Incorrect} \\
				\textbf{Evaluation of ours:} \color[RGB]{84, 180, 53}{Correct} \\
				\textbf{Human Evaluation:} \color{red}{Incorrect} \\
				\textbf{Evaluation Explanation of ours:} The user has asked about high-end Indian restaurants in town. The `restaurant-pricerange' has been correctly identified as `high' and the `restaurant-food' as `Indian'. ... The `area' slot has not been specified in the state of this turn, but since the user has not mentioned a specific area within the town, it should not be considered a missed domain-slot. ...
				Therefore, the state of this turn is complete as it captures all the relevant domain-slot-value pairs mentioned by the user in the current turn.\\
				\bottomrule
		\end{tabular}}
		\caption{Three examples that the evaluation results are different between ours and the MultiWOZ 2.4 annotation.}
		\label{case-different-human}
	\end{table*}
\section{Appendix: Prompt Template}
In this section, we introduce the prompt templates we designed in our paper, which are shown in Table \ref{prompt-accuracy} and \ref{prompt-completeness}.
\begin{table*}
	\centering
	\begin{tabular}{p{2\columnwidth}}
		\toprule
		\textbf{Prompt Template for Accuracy Evaluation:}\\
		\# Dialogue History:\\
		~~~~\{The dialogue history\}\
		
		\# Current Turn:\\
		\{\\
			~~~~"Agent": \{system\},\\
			~~~~"User": \{user\},\\
			~~~~"State of this turn": \{turn\_label\}\\
		\}\\
		
		\# Instruction:\\
		Your task is to evaluate all the domain-slot, value pairs in "State of this turn". The turn state is used to represent the user's requirements and confirmation. Note that domain information is just generic information. Constraints are determined by the slot, not the domain. Capitalization and the completeness is not a consideration.\\
		
		For each domain-slot, value pair, you should carefully assess whether the domain-slot, value pair is correct or not according to the current turn utterances and the history and gives some explanation. \\
		
		{\color{blue}{You should evaluate them one by one and finally output all the incorrect domain-slot, value, pairs.}}\\
		
		\# Output Format:\\
		Please output your analysis in JSON format as follows:\\
		\{\\
			~~~~"explanation": <a belief explanation of your judgement>,\\
			~~~~// incorrect domain-slot, value pairs in "State of this turn". Note that only domain-slot, value pairs in "State of this turn" should be evaluated.\\
			~~~~"incorrect\_domain\_slot": \{"domain-slot1": <value1>, ...\}\\
		\}\\
		\bottomrule
	\end{tabular}
	\caption{The prompt template for evaluating accuracy. Words in blue are the manually constructed reasoning path. }
	\label{prompt-accuracy}
\end{table*}
\begin{table*}
	\centering
	\begin{tabular}{p{2\columnwidth}}
		\toprule
		\textbf{Prompt Template for Completeness Evaluation:}\\
		\# Dialogue History:\\
			~~~~\{The dialogue history\}\

		\# Current Turn:\\
		\{\\
			~~~~"Agent": \{system\},\\
			~~~~"User": \{user\},\\
			~~~~"State of this turn": \{turn\_label\}\\
		\}\\
		
		Your task is to assess the completeness of the state of the current turn. The turn state is represented by a set of domain-slot, value pairs that represent only the state mentioned in the current turn utterances between the agent and the user, without considering the entire dialogue history.\\
		
		For completeness, you must determine if all relevant domain-slot-value pairs in the turn utterances have been captured in the turn state. Only domain-slot, value pairs that are new or have been updated in the current turn should be included. If a domain-slot-value pair has been mentioned previously in the dialogue and has not changed, it should not be considered a missed domain-slot.\\
		
		When the agent provides a recommendation and the user either confirms acceptance or requests more information, the user is considered to have accepted the recommendation. In this case, the turn state should include the details of the recommendation. However, information that the user has requested should not be included in the turn state, as the agent is expected to provide these details in the subsequent turn.If user express that he/she does not care about some domain-slot, the state should contain these domain-slots with the value "dontcare". Note that domain-slot that should be added but are not provided in the dialogue should not be considered as missed domain-slots.\\
		
		There are five domains that the AI agent supported, and their slots are listed in the following:\\
		1. Hotel: \{area, type, internet, parking, name, book day, price range, stars, book stay, book people\}\\
		...\\
		5. Train: \{book people, day, departure, destination, leave at\}\\
		~~\\
		Categorical slots and their possible values:\\
		1. Area: centre, east, south, west, north\\
		...\\
		10. Arrive by \& leave at: time in forms of “xx:xx” such as “13:00”\\
		
		~~\\
		
		Remember that domain-slot, value pairs mentioned in the dialogue history but not changed in the current turn should not be considered while evaluating.\\
		{\color{blue}{While evaluating, you should follow the following process: You should carefully read the turn utterances, while encountering a slot value, you should first read the history and determine whether it is already mentioned. If it has been mentioned, then determine whether it is updated in the current turn. Finally, if it is a domain-slot that should be tracked in the current turn, you should read the "state of this turn" and judge if it is missed or not.}}\\
		
		\# Output Format:\\
		Please output your analysis in JSON format as follows:\\
		\{\\
				~~~~"explanation": <the process of your step by step thinking>,\\
				~~~~// missed domain-slot, value pairs in "State of this turn". Note that domain-slot, value pairs not mentioned in the current turn utterances should not be considered and those value has not been expressed should not be included either.\\
				~~~~"missed\_domain\_slot": \{"domain-slot1": <its corresponding value>, ...\}\\
		\}\\
		\bottomrule
	\end{tabular}
	\caption{The prompt template for evaluating completeness. Words in blue are the manually constructed reasoning path. }
	\label{prompt-completeness}
\end{table*}

\end{document}